%% file: ms.tex
\documentclass[a4paper, 10pt, conference]{ieeeconf}      

\IEEEoverridecommandlockouts                              

\usepackage{xcolor}
\usepackage{soul}
\usepackage[utf8]{inputenc}
\usepackage[small]{caption}
\usepackage{booktabs}
\usepackage{graphics} 
\usepackage{tabularx}
\usepackage{amsmath} 
\usepackage{amssymb}  
\usepackage{mathrsfs}
\usepackage{subcaption}
\usepackage{tikz}
\usetikzlibrary{shapes,arrows, automata}
\usepackage{pgfplots}
\usepackage[per-mode=symbol]{siunitx}
\DeclareSIUnit{\mph}{mph}
\usepackage{url}
\usepackage[noadjust]{cite}
\usepackage{numprint}
\usepackage{bm}
\newcommand{\vect}[1]{\boldsymbol{\mathbf{#1}}}

\pgfplotsset{compat=newest}
\pgfplotsset{every axis/.append style={
	font=\LARGE}
}

\pgfplotsset{every axis legend/.append style={legend cell align=left}}

\def\man#1;{%
    \begin{scope}[shift={#1}]
        \fill [rounded corners=1.5] (0,0.4) -- (0,0.8) -- (0.4,0.8) -- (0.4,0.4) --
            (0.325,0.4) -- (0.325,0.7) -- (0.3,0.7) -- (0.3,0) -- (0.225,0) --
            (0.225,0.4) -- (0.175,0.4) -- (0.175,0) -- (0.1,0) -- (0.1,0.7) --
            (0.075,0.7) -- (0.075,0.4) -- cycle;
        \fill (0.2,0.9) circle (0.1);
    \end{scope}}
    
\newcommand{\argmax}{\operatornamewithlimits{arg\,max}}

\usetikzlibrary{arrows.meta, calc, shapes}
\tikzset{%
  >={Latex[width=2mm,length=2mm]},
            base/.style = {rectangle, rounded corners, draw=black,
                           minimum width=1cm, minimum height=1cm,
                           text centered, font=\sffamily},
            simulator/.style = {base, fill=green!30, minimum width=4cm},
            solver/.style = {base, fill=red!30},
            reward/.style = {base, minimum height=1.5cm},
            module/.style = {base, minimum width=2.5cm, minimum height=1.5cm, fill=blue!30},
            module2/.style = {base, minimum width=2.5cm, minimum height=1.5cm, fill=white},
            network/.style = {base, minimum width=2.5cm, minimum height=1.5cm, fill=white},
            state/.style = {base, minimum width=0.5cm, minimum height=1.0cm, fill=white},
            io/.style = {base, minimum width=0.5cm, minimum height=1.0cm, fill=white},
}
\pgfdeclarelayer{background}
\pgfdeclarelayer{foreground}
\pgfsetlayers{background,main,foreground}

\newif\ifcuboidshade
\newif\ifcuboidemphedge

\tikzset{
  cuboid/.is family,
  cuboid,
  shiftx/.initial=0,
  shifty/.initial=0,
  dimx/.initial=3,
  dimy/.initial=3,
  dimz/.initial=3,
  scale/.initial=1,
  densityx/.initial=1,
  densityy/.initial=1,
  densityz/.initial=1,
  rotation/.initial=0,
  anglex/.initial=0,
  angley/.initial=90,
  anglez/.initial=225,
  scalex/.initial=1,
  scaley/.initial=1,
  scalez/.initial=0.5,
  front/.style={draw=black,fill=white},
  top/.style={draw=black,fill=white},
  right/.style={draw=black,fill=white},
  shade/.is if=cuboidshade,
  shadecolordark/.initial=black,
  shadecolorlight/.initial=white,
  shadeopacity/.initial=0.15,
  shadesamples/.initial=16,
  emphedge/.is if=cuboidemphedge,
  emphstyle/.style={thick},
}

\newcommand{\tikzcuboidkey}[1]{\pgfkeysvalueof{/tikz/cuboid/#1}}

\newcommand{\tikzcuboid}[1]{
    \tikzset{cuboid,#1} 
  \pgfmathsetlengthmacro{\vectorxx}{\tikzcuboidkey{scalex}*cos(\tikzcuboidkey{anglex})*28.452756}
  \pgfmathsetlengthmacro{\vectorxy}{\tikzcuboidkey{scalex}*sin(\tikzcuboidkey{anglex})*28.452756}
  \pgfmathsetlengthmacro{\vectoryx}{\tikzcuboidkey{scaley}*cos(\tikzcuboidkey{angley})*28.452756}
  \pgfmathsetlengthmacro{\vectoryy}{\tikzcuboidkey{scaley}*sin(\tikzcuboidkey{angley})*28.452756}
  \pgfmathsetlengthmacro{\vectorzx}{\tikzcuboidkey{scalez}*cos(\tikzcuboidkey{anglez})*28.452756}
  \pgfmathsetlengthmacro{\vectorzy}{\tikzcuboidkey{scalez}*sin(\tikzcuboidkey{anglez})*28.452756}
  \begin{scope}[xshift=\tikzcuboidkey{shiftx}, yshift=\tikzcuboidkey{shifty}, scale=\tikzcuboidkey{scale}, rotate=\tikzcuboidkey{rotation}, x={(\vectorxx,\vectorxy)}, y={(\vectoryx,\vectoryy)}, z={(\vectorzx,\vectorzy)}]
    \pgfmathsetmacro{\steppingx}{1/\tikzcuboidkey{densityx}}
  \pgfmathsetmacro{\steppingy}{1/\tikzcuboidkey{densityy}}
  \pgfmathsetmacro{\steppingz}{1/\tikzcuboidkey{densityz}}
  \newcommand{\dimx}{\tikzcuboidkey{dimx}}
  \newcommand{\dimy}{\tikzcuboidkey{dimy}}
  \newcommand{\dimz}{\tikzcuboidkey{dimz}}
  \pgfmathsetmacro{\secondx}{2*\steppingx}
  \pgfmathsetmacro{\secondy}{2*\steppingy}
  \pgfmathsetmacro{\secondz}{2*\steppingz}
  \foreach \x in {\steppingx,\secondx,...,\dimx}
  { \foreach \y in {\steppingy,\secondy,...,\dimy}
    {   \pgfmathsetmacro{\lowx}{(\x-\steppingx)}
      \pgfmathsetmacro{\lowy}{(\y-\steppingy)}
      \filldraw[cuboid/front] (\lowx,\lowy,\dimz) -- (\lowx,\y,\dimz) -- (\x,\y,\dimz) -- (\x,\lowy,\dimz) -- cycle;
    }
    }
  \foreach \x in {\steppingx,\secondx,...,\dimx}
  { \foreach \z in {\steppingz,\secondz,...,\dimz}
    {   \pgfmathsetmacro{\lowx}{(\x-\steppingx)}
      \pgfmathsetmacro{\lowz}{(\z-\steppingz)}
      \filldraw[cuboid/top] (\lowx,\dimy,\lowz) -- (\lowx,\dimy,\z) -- (\x,\dimy,\z) -- (\x,\dimy,\lowz) -- cycle;
        }
    }
    \foreach \y in {\steppingy,\secondy,...,\dimy}
  { \foreach \z in {\steppingz,\secondz,...,\dimz}
    {   \pgfmathsetmacro{\lowy}{(\y-\steppingy)}
      \pgfmathsetmacro{\lowz}{(\z-\steppingz)}
      \filldraw[cuboid/right] (\dimx,\lowy,\lowz) -- (\dimx,\lowy,\z) -- (\dimx,\y,\z) -- (\dimx,\y,\lowz) -- cycle;
    }
  }
  \ifcuboidemphedge
    \draw[cuboid/emphstyle] (0,\dimy,0) -- (\dimx,\dimy,0) -- (\dimx,\dimy,\dimz) -- (0,\dimy,\dimz) -- cycle;%
    \draw[cuboid/emphstyle] (0,\dimy,\dimz) -- (0,0,\dimz) -- (\dimx,0,\dimz) -- (\dimx,\dimy,\dimz);%
    \draw[cuboid/emphstyle] (\dimx,\dimy,0) -- (\dimx,0,0) -- (\dimx,0,\dimz);%
    \fi

    \ifcuboidshade
    \pgfmathsetmacro{\cstepx}{\dimx/\tikzcuboidkey{shadesamples}}
    \pgfmathsetmacro{\cstepy}{\dimy/\tikzcuboidkey{shadesamples}}
    \pgfmathsetmacro{\cstepz}{\dimz/\tikzcuboidkey{shadesamples}}
    \foreach \s in {1,...,\tikzcuboidkey{shadesamples}}
    {   \pgfmathsetmacro{\lows}{\s-1}
        \pgfmathsetmacro{\cpercent}{(\lows)/(\tikzcuboidkey{shadesamples}-1)*100}
        \fill[opacity=\tikzcuboidkey{shadeopacity},color=\tikzcuboidkey{shadecolorlight}!\cpercent!\tikzcuboidkey{shadecolordark}] (0,\s*\cstepy,\dimz) -- (\s*\cstepx,\s*\cstepy,\dimz) -- (\s*\cstepx,0,\dimz) -- (\lows*\cstepx,0,\dimz) -- (\lows*\cstepx,\lows*\cstepy,\dimz) -- (0,\lows*\cstepy,\dimz) -- cycle;
        \fill[opacity=\tikzcuboidkey{shadeopacity},color=\tikzcuboidkey{shadecolorlight}!\cpercent!\tikzcuboidkey{shadecolordark}] (0,\dimy,\s*\cstepz) -- (\s*\cstepx,\dimy,\s*\cstepz) -- (\s*\cstepx,\dimy,0) -- (\lows*\cstepx,\dimy,0) -- (\lows*\cstepx,\dimy,\lows*\cstepz) -- (0,\dimy,\lows*\cstepz) -- cycle;
        \fill[opacity=\tikzcuboidkey{shadeopacity},color=\tikzcuboidkey{shadecolorlight}!\cpercent!\tikzcuboidkey{shadecolordark}] (\dimx,0,\s*\cstepz) -- (\dimx,\s*\cstepy,\s*\cstepz) -- (\dimx,\s*\cstepy,0) -- (\dimx,\lows*\cstepy,0) -- (\dimx,\lows*\cstepy,\lows*\cstepz) -- (\dimx,0,\lows*\cstepz) -- cycle;
    }
    \fi 

  \end{scope}
}

\makeatother

\makeatletter
\let\NAT@parse\undefined
\makeatother
\usepackage{hyperref} 
\usepackage{cleveref}

\title{\Large \bf Efficient Autonomy Validation in Simulation\\ with Adaptive Stress Testing}
\author{Mark Koren$^{1}$ and Mykel J. Kochenderfer$^{1}$
\thanks{$^{1}$Mark Koren and Mykel J. Kochenderfer are with Aeronautics and Astronautics, Stanford University, Stanford, CA 94305, USA
        {\tt\small \{mkoren, mykel\}@stanford.edu}}
}

\begin{document}

\maketitle
\thispagestyle{empty}
\pagestyle{empty}

\begin{abstract}  
During the development of autonomous systems such as driverless cars, it is important to characterize the scenarios that are most likely to result in failure.
Adaptive Stress Testing (AST) provides a way to search for the most-likely failure scenario as a Markov decision process (MDP).
Our previous work used a deep reinforcement learning (DRL) solver to identify likely failure scenarios.
However, the solver's use of a feed-forward neural network with a discretized space of possible initial conditions poses two major problems.
First, the system is not treated as a black box, in that it requires analyzing the internal state of the system, which leads to considerable implementation complexities.
Second, in order to simulate realistic settings, a new instance of the solver needs to be run for each initial condition.
Running a new solver for each initial condition not only significantly increases the computational complexity, but also disregards the underlying relationship between similar initial conditions.
We provide a solution to both problems by employing a recurrent neural network that takes a set of initial conditions from a continuous space as input.
This approach enables robust and efficient detection of failures because the solution generalizes across the entire space of initial conditions.
By simulating an instance where an autonomous car drives while a pedestrian is crossing a road, we demonstrate the solver is now capable of finding solutions for problems that would have previously been intractable.
\end{abstract}

\section{INTRODUCTION}\label{sec:intro}

Simulation can offer an inexpensive complement to field-testing for evaluating the safety of autonomous vehicles~\cite{Agaram2016, Koopman2016, Kalra2016}. 
Such simulations can run faster than real-time and can more easily probe safety critical scenarios that cannot be obtained in real-world environments due to the rarity of events, cost incurred with failures, and ethical considerations.
However, the space of edge-cases that can cause the autonomous vehicle to fail is vast~\cite{Koopman2018}. 

\begin{figure}[t]
    \setlength\belowcaptionskip{-0.75\baselineskip}	
	\centering
    \vspace*{0.25cm}
    \centering
    \includegraphics[width=0.70\columnwidth]{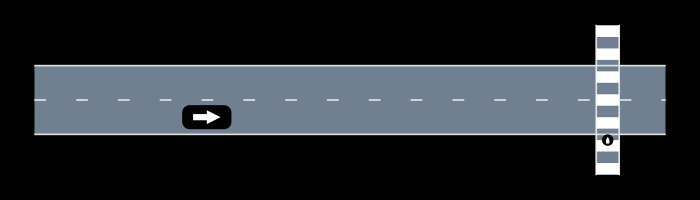}
    \caption{Layout of a running example. A car approaches a crosswalk on a neighborhood road with one lane in each direction. A pedestrian is attempting to cross the street at the crosswalk.}
	\label{fig:scenario1} 
\end{figure}

Consider a pedestrian crossing a neighborhood road at a crosswalk, a problem we use as a running example and which is shown in \Cref{fig:scenario1}. A naive approach could be to assume the pedestrian follows  a straight line trajectory across the road. The scenario could then be simulated thousands of times with different pedestrian velocities. While this approach may be tractable, the computational savings come at the expense of safety. In reality, pedestrians do not only follow a straight line. Perhaps certain pedestrian paths take them into a sensor blind-spot, or elicit unsafe behavior from the test vehicle. Even unlikely collisions are significant due to the large amount of miles an autonomous fleet will drive. It is not sufficient to assume other actors will always follow sensible trajectories. Unfortunately, modeling the behavior of other actors leads to an exponential explosion in possible scenarios and failures. Identifying critical scenarios through a brute-force search would be intractable due to the dimensionality of the search-space. Instead, researchers are focusing on adaptive methods for adversarially generating critical test scenarios in simulation~\cite{mullins2018adaptive, o2018scalable, tuncali2018simulation, li2016intelligence}.

Adaptive Stress Testing (AST) provides a framework for finding the most-likely failure scenario of a system in simulation~\cite{lee2015adaptive}.
Knowing the most-likely failure is useful for both the development and the safety validation of an autonomous vehicle.
In AST, the problem of finding the most-likely failure of a system is formulated as a Markov decision process (MDP). Failures can be found using reinforcement learning techniques. The process of solving an AST problem is shown in \Cref{fig:ASTStruct}.
At each time-step, the solver provides \textit{environment actions} that control the simulator.
The simulator reports when a failure occurs and outputs the likelihood of the environment actions.
Reinforcement learning techniques can be used to solve the MDP, with the reward function depending on the likelihood of actions taken and whether a failure was found.
We recently introduced a deep reinforcement learning (DRL) solver that was able to find failures in an example autonomous vehicle scenario more efficiently than an existing Monte Carlo tree search (MCTS) solver~\cite{Koren}.
However, there are two major limitations that make using the solver more challenging: the solver's dependence on the simulation state and requirement to be run from a single initial condition.

The primary limitation of the DRL solver is its dependence on observing the simulation state, the collection of internal state variables that fully define the simulation.
Many high-fidelity simulators are large, complicated software suites.
Consequently, exposing the simulation state may be non-trivial; therefore it is useful to treat the simulator as a black box. 
In the current formulation of AST, the method is able to treat the simulation as a black box by using the history of actions taken as a substitute for the current state~\cite{lee2015adaptive}.
The system under test (SUT) must therefore be deterministic with respect to the environment actions, so that an exact mapping from the history of actions to simulation state exists.
The DRL solver represents the AST policy as a feed-forward neural network, where the simulation state is used as input. 
It would be advantageous to use the history of actions as input, but the solver architecture is not optimal for such a representation. 
Instead, an  architecture designed for sequential data would be preferable. 

\begin{figure}[t]
    \setlength\belowcaptionskip{-0.75\baselineskip}	
	\centering
    \scalebox{0.8}{\input{Images/ASTStruct.tex}}
    \caption{The AST methodology. The simulator is treated as a black box. The solver optimizes a reward based on transition likelihood and whether an event has occurred.}
	\label{fig:ASTStruct} 
\end{figure}
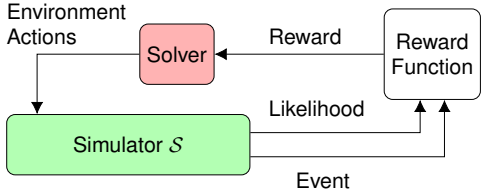
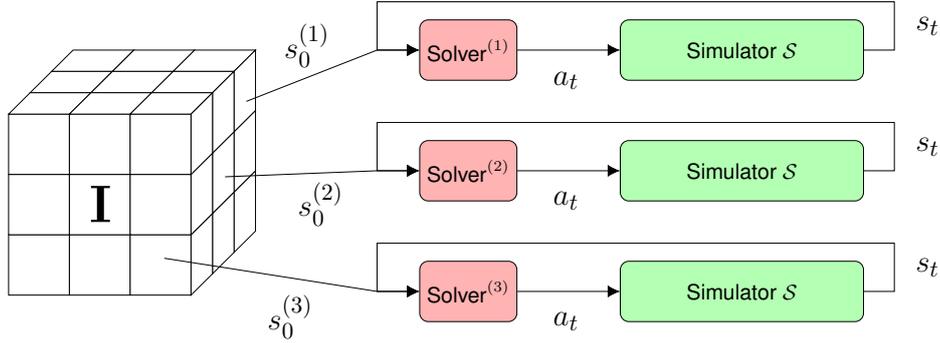
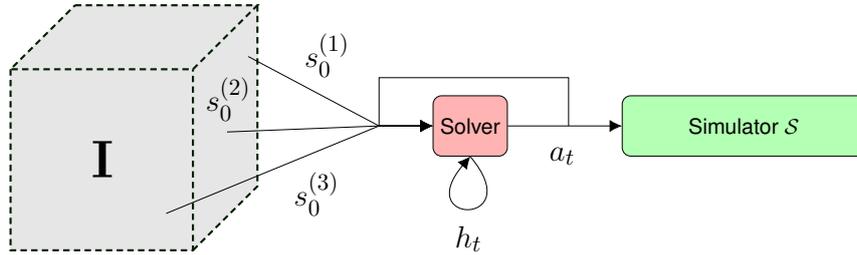
\begin{figure*}[t!]
	\centering
    \begin{subfigure}[t]{1.0\textwidth}
        \centering
        \scalebox{0.8}{\input{Images/arch.tex}}
        \caption{The previous version of AST running over a space of initial conditions $\mathbf{I}$. The space must be discretized, and then each initial condition $s_0$ requires a separate instance of the DRL solver and the simulator. In addition, the solver requires the next simulator state $s_t$ at each time-step.}
        \label{fig:arch_old}
    \end{subfigure}
    \begin{subfigure}[t]{1.0\textwidth}
        \centering
        \scalebox{0.8}{\input{Images/newArch.tex}}
        \caption{The new version of AST running over a space of initial conditions $\mathbf{I}$. The continuous space is sampled at random at the start of each trajectory, therefore only one instance of the solver is needed. The solver does not need access to the simulation state because it is maintaining a hidden state $h_t$ at each time-step. The solver instead uses the previous action $a_t$.}
        \label{fig:arch_new}
    \end{subfigure}
    \setlength\belowcaptionskip{-0.75\baselineskip}	
    \caption{Contrasting the new and old AST architectures. The new solver uses a recurrent architecture and is able to generalize across a continuous space of initial conditions with a single trained instance. These improvements allow AST to be used on problems that would have previously been intractable.}
	\label{fig:arch_comparison} 
\end{figure*}

The second limitation of the DRL solver is that it can only be trained for a single initial condition. 
Consider again the pedestrian at the crosswalk.
Simulating this scenario with the car starting \SI{30}{\meter} away from the crosswalk could have a significantly different outcome than if the car started \SI{40}{\meter} away.
When validating an autonomous vehicle, we are interested in both of these instantiations, as well as the range in-between.
All of the possible initial conditions comprise a class of scenarios, and different instantiations could lead to different failure modes. 
An ideal validation method would therefore be able to cover the entire initial condition space. 
Unfortunately, we currently would have to discretize the space of initial conditions and train a new DRL solver for each bin, as shown in \Cref{fig:arch_old}.
Even on small problems, the space of initial conditions can be high-dimensional, making a discretized representation intractable.
Instead, we would like to train a single DRL solver that can take any initial condition in the continuum of our defined space.

In order to effectively use AST while validating complex autonomous vehicles, this paper extends the DRL solver by changing the policy architecture to a recurrent neural network (RNN), which has two advantages:
\begin{enumerate}
    \item RNNs are designed for sequential tasks, therefore the simulation state is no longer needed as input.
    The RNN takes the previous action as input, and uses it to internally maintain a hidden state.
    This is analogous to using the history of actions as the current state. 
    \item Specific types of RNNs have shown the ability to learn temporal patterns.
    This is essential when working with different initial conditions, since trajectories could end up in similar states that have different expected values due to reaching the state different times.
    Therefore, we are able to add the initial state as input to the network, and the network will learn to generalize across the space of initial conditions.
\end{enumerate}
The new architecture therefore addresses the two major limitations of the old DRL solver.
We will demonstrate the improvements by roughly discretizing the space of initial conditions and comparing the performance of the new architecture against a MCTS solver and the current DRL solver. 
Generalization will then be demonstrated by letting the new architecture sample from the space of initial conditions at train time. 
Improving the DRL solver will make autonomous vehicle validation in simulation more tractable, leading to vehicles that are more reliable and robust.

\section{BACKGROUND} \label{sec:bg}
Adaptive stress testing formulates the problem of finding the most-likely decision as an MDP.
Two methods are used to solve this problem, MCTS and DRL.

\subsection{Markov Decision Process} \label{sec:bg_mdp}

Markov decision processes (MDPs) are a framework for formulating sequential decision making problems~\cite{DMU}. In an MDP an agent chooses an action $a$ in state $s$. The agent receives a reward according to the reward function $R(s,a)$. The agent transitions to the next state $s'$ according to the transition probability function $T(s'\mid s,a)$. According to the Markov assumption, the transition only depends on the current state and action. Neither the transition nor reward functions need to be deterministic. In some cases, the transition or reward functions may not be known. 

The solution to an MDP is represented by a policy $\pi(s)$ that specifies the optimal action to take in a given state. An optimal action maximizes the expected utility, which can be found recursively:
\begin{equation}
V^{\pi} \left(s\right) = R\big(s, \pi\left(s\right)\big) + \gamma  \sum_{s'}T\big(s'\mid s,\pi \left(s\right)\big)V^{\pi}\left(s'\right)
\end{equation}
where $\gamma$ is the discount factor that controls the weight of future rewards. Large MDPs may need to be solved approximately. Two examples of reinforcement learning techniques for finding the approximate solution to an MDP are Monte Carlo tree search and deep reinforcement learning.
$$$$
\subsection{Monte Carlo Tree Search} \label{sec:bg_mcts}
Monte Carlo tree search (MCTS)~\cite{MCTSUCT} is an online sampling-based reinforcement learning method that has performed well on large MDPs~\cite{MCTS_GO}. MCTS builds and maintains a tree where the nodes represent states or actions in the MDP. While executing from states in the tree, MCTS chooses the action that maximizes   
\begin{equation}
\label{eq:UCB}
a\gets \argmax_a Q(s,a)+c\sqrt{\frac{\log(N(s))}{N(s,a)}}
\end{equation}
where $Q(s,a)$ is the expected value of a state-action pair, $N(s)$ and $N(s,a)$ are the number of times a state and a state-action pair have been visited, respectively, and $c$ is a parameter that controls exploration. When a new state is encountered, $Q(s,a)$ and $N(s,a)$ are initialized for all of the available actions, and the state is added to the tree. Then, $Q(s,a)$ is updated by executing rollouts to a specified depth and returning the expected value. The algorithm is run until a stopping criterion is met. This paper uses a variation of MCTS with double progressive widening~\cite{MCTSDPW} to limit the branching of the tree, which leads to better performance on problems with large or continuous action spaces.

\subsection{Deep Reinforcement Learning} \label{sec:bg_drl}
Deep reinforcement learning (DRL) represents a policy as a neural network (NN) parameterized by $\theta$. Recurrent neural networks (RNN) are a family of NNs designed to handle sequential inputs. Recurrent neural networks maintain a hidden state, which propagates information through time. RNNs factor historical information into their output through the hidden state. The network maintains a set of weights for both the hidden state and the output. While RNNs traditionally can be difficult to train due to the exploding gradient problem, long-short term memory (LSTM) layers fix this by introducing gated self-loops which enforce constant error flow~\cite{Hochreiter1997}. 

Trust Region Policy Optimization (TRPO) is a gradient-based method for improving the policy~\cite{Schulman2015a}. TRPO generally gives monotonic increases in policy performance by constraining the KL divergence between policy steps. The policy gradient can be obtained using generalized advantage estimation (GAE)~\cite{Schulman2015}, a method for estimating policy gradients from batches of simulation trajectories. 

\section{Methodology}\label{sec:ast}
When validating autonomous systems, stress testing is the process of eliciting failures to evaluate the robustness of the system. This section outlines the process involved in using AST to find the most-likely failure, based on the material presented above. We also explain the changes made to the DRL solver to improve the performance and add a new capability.
\subsection{Adaptive Stress Testing} \label{sec:meth_ast}
Adaptive stress testing formulates finding the most-likely failure of a system as a sequential decision-making problem. Given a simulator $\mathcal{S}$ and a subset of the state space $E$ where the events of interest (e.g. a collision) occurs, we want to find the most-likely trajectory $s_0, \ldots, s_t$ that ends in our subset $E$. Given $(\mathcal{S}, E)$, the formal problem is
\begin{equation*}
\begin{aligned}
& \underset{a_0, \ldots, a_t}{\text{maximize}}
& & P(s_0, a_0, \ldots,s_t, a_t) \\
& \text{subject to}
& & s_t \in E
\end{aligned}
\end{equation*}
where $P(s_0, a_0, \ldots,s_t, a_t)$ is the probability of a trajectory in simulator $\mathcal{S}$ and $s_t = f(a_t, s_{t-1})$.


AST requires the following three functions to interact with the simulator:
\begin{itemize}
\item \textsc{Initialize}$(\mathcal{S}, s_0)$: Resets $\mathcal{S}$ to a given initial state $s_0$.
\item \textsc{Step}$(\mathcal{S}, E, a)$: Steps the simulation in time by drawing the next state $s'$ after taking action $a$. The function returns the probability of the transition and an indicator whether $s'$ is in $E$ or not.
\item \textsc{IsTerminal}$(\mathcal{S}, E)$: Returns true if the current state of the simulation is in $E$, or if the horizon of the simulation $T$ has been reached. 
\end{itemize}
Unlike previous formulations, the \textsc{Initialize} function now accepts an initial state. The purpose of this change is to output a policy that can generalize to different scenario instantiations.

\subsection{Recurrent Deep Reinforcement Learning Solver} \label{sec:drl_solver}
We previously added a new deep reinforcement learning (DRL) solver to AST~\cite{Koren}. The solver is interchangeable with the commonly-used MCTS solver. The previous implementation required the simulation state as input, which was an undesirable relaxation of the black-box simulator assumption. Treating the simulation as a black box allows easier implementation for complicated or third-party simulators, for which the simulator's internal state may not be accessible. As such, we have redesigned the DRL solver to meet the black-box assumption.

The AST agent must control all stochasticity in the simulation, therefore transitions are deterministic with respect to the AST agent’s actions. Because the SUT updates are deterministic, the history of actions and the initial state are sufficient to represent the current state. Consequently, the simulator is allowed to be non-markovian. Replacing the simulation state with the history of actions also fulfills the black-box assumption, because the simulation state is no longer needed as input. The previous DRL solver, referred to hereafter as the MLPDRL solver, used a Gaussian multi-layer perceptron (MLP) architecture, which does not work well with this state representation. 

\begin{figure}[htbp]
    \setlength\belowcaptionskip{-0.75\baselineskip}	
	\centering
    \scalebox{0.8}{\input{Images/NNArch.tex}}
    \caption{The DRL solver architecture. The recurrent neural network structure takes the hidden state from the previous step, $h_t$, and outputs the hidden state for the next step $h_{t+1}$. The input $x_t$ is a concatenation of the initial state $s_0$ and the previous action $a_t$. The output $y_t$ is the mean $\mu_{t+1}$ and standard deviation  $\Sigma_{t+1}$ of the next action.}
	\label{fig:NNArch} 
\end{figure}
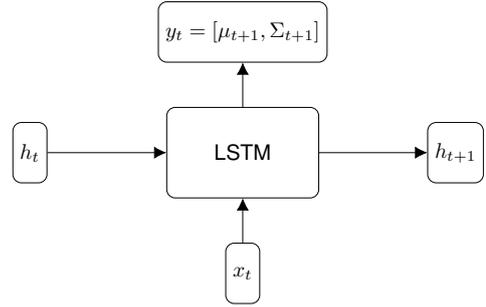

Instead of a Gaussian MLP, the policy is now represented by a recurrent neural net (RNN), using long-short term memory (LSTM) layers. The network architecture is shown in~\Cref{fig:NNArch}. An RNN is able to train on a sequence of inputs while maintaining a hidden state, which is analogous to using a history of previous actions as the current state. The output of the policy is a mean action vector for a multivariate Gaussian distribution. The diagonal covariance matrix is independent of state and trained separately~\cite{Schulman2015a}. The only input to the network is the previous action, hence $x_t = a_{t-1}$. While the simulation state is no longer needed, this solver can only be run from a single initial condition, therefore we will refer to it as the discrete recurrent deep reinforcement learning (DRDRL) solver.

\subsection{Continuous Scenario Generalization} \label{sec:meth_scen}
Previous work with AST provided a trajectory from a discrete initial condition. As discussed earlier, engineers are often concerned with a scenario that starts from a space of initial conditions. Even using a coarse grid discretization, the number of possible initial conditions is still exponential in the dimension of the initial condition space. Despite the increased efficiency of AST, running from a large number of initial conditions for a single class of scenarios would take an impractical amount of compute time. Instead, we would like to have a single solver that can find a likely failure trajectory from any initial condition.

By modifying the \textsc{Initialize} function to accept the initial state, we hypothesize that AST can learn a policy that generalizes to the entire space of initial conditions. Our hypothesis arises because autonomous vehicles test cases are run on models of the real world. Consequently, light deviations in position and noise should produce similar policies. Therefore, training AST across the space of initial conditions should be far more sample efficient than running AST for individual instantiations. The architecture is therefore modified so the input at each time-step is the concatenation of the previous action and the initial condition, hence $x_t = [a_{t-1}, s_0]$. During training, each rollout starts from a randomly sampled intial condition. We will refer to this architecture as the generalized recurrent deep reinforcement learning (GRDRL) solver.
\section{Experiments}\label{sec:setup}
This section outlines the problem used in simulation to test AST,  the hyper-parameters of the DRL solver, and the reward structure. For bench-marking purposes, we follow the experiment setup---simulation, pedestrian models, and SUT model---proposed in our previous work~\cite{Koren}. The problem has a 5-dimensional state-space and a 6-dimensional action space, and is run for up to 50 time-steps. 
\subsection{Problem Formulation}

Our experiment simulates a common neighborhood road driving scenario, shown in~\Cref{fig:scenario1}. The road has one lane in each direction. A pedestrian crosses at a marked crosswalk, from south to north. The $y$ origin is at the center of the crosswalk, and the $x$ origin is where the crosswalk meets the side of the road. The speed limit is \SI{25}{\mph}, which is \SI{11.17}{\meter\per\second}. 

The inputs to the GRDRL solver include the initial state $\vect{s}_0 =[\vect{s}_{0, ped}, s_{0, car}, v_{0, ped}, v_{0, car}]$ where
\begin{itemize}
\item $\vect{s}_{0, ped}$ is the initial $x$, $y$ location of the pedestrian,
\item $s_{0, car}$ is the initial $x$ position of the car,
\item $v_{0, ped}$ is the initial $y$ velocity of the pedestrian, and
\item $v_{0, car}$ is the initial $x$ velocity of the car.
\end{itemize}

Initial conditions are drawn from a continuous uniform distribution, with the supports shown in \Cref{table:variables}. Trajectory rollouts are instantiated by randomly sampling an initial condition from the parameter ranges. 
\begin{table}[h]
    \caption{The initial condition space. Initial conditions are drawn from a continuous uniform distribution defined by the supports below.}
    \label{table:variables}
    \begin{center}
        \input{Images/table_variables.tex}
    \end{center}
    \vspace{-7mm}
\end{table}
\subsection{Modified Reward Function}
AST penalizes each step by the likelihood of the environment actions, as shown in~\Cref{eq:1}. Unlikely actions have a higher cost, so the solver is incentivized to take likelier actions, and therefore find likelier failures. The Mahalanobis distance~\cite{mahalanobis1936generalised} is used as a proxy for the likelihood of an action. The Mahalanobis distance is a measure of distance from the mean generalized for multivariate continuous distributions. The penalty for failing to find a collision is controlled by $\alpha$ and $\beta$. The penalty at the end of a no-collision case  is scaled by the distance ($\textsc{dist}$) between the pedestrian and the vehicle. The penalty encourages the pedestrian to end early trials closer to the vehicle and leads to faster convergence. We use $\alpha = \SI{-1e5}{}$ and $\beta = \SI{-1e4}{}$. The reward function is modified from the previous version of AST ~\cite{lee2015adaptive} as follows:
\begin{equation}
\label{eq:1}
R\left(s\right) = \left\{
        \begin{array}{ll}
            0, &  s \in E \\
            -\alpha - \beta\times\textsc{dist}\left(\vect p_v,\vect p_p\right), &  s \notin E, t\geq T \\
            - M\left(a, \mu_a, \Sigma_a\mid s\right),   &  s \notin E, t < T
        \end{array}
    \right.
\end{equation}
where $ M(a, \mu_a, \Sigma_a\mid s)$ is the Mahalanobis distance between the action $a$ and the expected action $\mu_a$ given the covariance matrix $\Sigma_a$ in the current state $s$. The distance between the vehicle position $\vect p_v$ and the closest pedestrian position $\vect p_p$ is given by the function $\textsc{dist}(\vect p_v,\vect p_p)$.
\subsection{Solvers}
The DRL solver uses the new recurrent architecture shown in~\Cref{fig:NNArch}. The hidden layer size is 64. Training was done with a batch size of \num{5e5} time-steps. The maximum trajectory length is \num{50}, hence each batch has \num{1000} trajectories. The optimizer used a step size of \num{0.1}, and a discount factor of \num{0.99}. The DRL approach was implemented using rllab~\cite{duan2016benchmarking}.
\section{RESULTS}\label{sec:results}
This section shows the performance of the new solvers on our running example. First, the solvers ability to train on the problem are compared to each other. Both solvers are then compared to baselines to show their improvement.
\subsection{Overall Performance}
\begin{figure}[h!]
	\centering
    \scalebox{0.75}{\input{Images/test.tex}}
    \caption{The Mahalanobis distance of the most-likely failure found at each iteration for both architectures. The conservative discrete architecture runs each of the discrete solvers in sequential order. The optimistic discrete architecture runs each of the discrete solvers in a single batch.}
	\label{fig:overall} 
\end{figure}
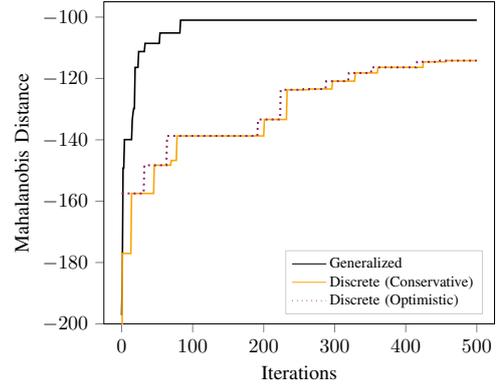
The goal of AST is to understand failure modes by returning the most-likely failure. An advantage of the new architecture is being able to search for the most-likely failure from a space of initial conditions while training a single network. \Cref{fig:overall} demonstrates these benefits by showing the cumulative maximum reward found by the DRDRL and GRDRL solvers at each iteration. There are two estimates shown for the DRDRL architecture:
\begin{itemize}
    \item Sequential: Each discrete AST solver is run sequentially. The naive approach serves as a lower bound on the performance of the discrete architecture.
    \item Batch: The AST solvers are updated as a batch. Each batch is assumed to still take 32 iterations, but the best reward of the best solver is known after each update.
\end{itemize}

The generalized architecture outperforms the discrete architecture at every iteration. The generalized version finds a collision sooner and converges to a solution after about 100 iterations, whereas the discrete architecture is still improving after 500 iterations. Furthermore, the generalized version is able to find a trajectory that has a net Mahalanobis distance of \num{-100.9901809119}. In contrast, the discrete version's most-likely solution was \num{-114.18531858492}. Over the entire space of initial conditions, running the generalized architecture is more accurate in far fewer iterations than running the discrete architecture at discrete points. 

\subsection{Comparison to Baselines}
\begin{table*}[h]
    \caption{The aggregate results of the DRDRL and GRDRL solvers, as well as the MCTS and MLPDRL solvers as baselines, on an autonomous driving scenario with a 5-dimensional initial condition space. Despite not having access to the simulator's internal state, the DRDRL is competitive with both baselines. However, the GRDRL solver demonstrates a significant improvement over the other three solvers. }
     \vspace{-1mm}
    \label{table:summary}
    \begin{center}
        \input{Images/table_summary.tex}
    \end{center}
    \vspace{-5mm}
\end{table*}


\Cref{table:summary} shows the aggregate results of the new architectures as well as two baselines: the old multi-layer perceptron architecture and a Monte Carlo tree search (MCTS) solver. The data was generated by dividing the 5-dimensional initial condition space into 2 bins per dimension, which resulted in 32 bins. Such a rough discretization is unsafe, but the number of bins is equal to $b^5$, where $b$ is the number of bins per dimension. Using 3 bins per dimension, which is hardly much safer, would result in training 243 instances of AST. Even on a toy problem, running AST for a safe number of discrete points is intractable. However, to demonstrate the performance benefits of the GRDRL architecture, we ran the MCTS, MLPDRL, and DRDRL solvers at the center-point each of the 32 bins. The GRDRL solver was trained on the entire space of initial conditions, and evaluated in two ways: 1) by executing the GRDRL solver's policy from the same 32 center-points of the other solvers were tested at, referred to as point evaluation, and 2) by sampling from each bin in the initial condition space and keeping the best GRDRL solution, referred to as bin evaluation. The performance of the DRDRL and GRDRL solvers are discussed below.

\subsubsection{Discrete Recurrent Deep Reinforcement Learning Solver}
The DRDRL solver performs similarly to both baselines. The DRDRL solver's average collision over the 32 bins was slightly worse than the MLPDRL solver and significantly worse than the MCTS solver. However, the DRDRL solver found crashes in many more bins than the MCTS solver, and found the most-likely collision of the three solvers. Finding the most-likely collision was the primary goal of the three solvers, therefore the DRDRL solver performed better than both baselines, despite not having access to the simulation state like the MLPDRL solver.

\subsubsection{Generalized Recurrent Deep Reinforcement Learning Solver}
The GRDRL solver far outperforms both baselines, as well as the DRDRL solver. When evaluating over the entire bin, the GRDRL solver found collisions in every single bin, and had by far the best average and maximum collision rewards. The maximum reward in particular demonstrates both the strength and necessity of the new solver architecture. The most-likely collision was not at one of the 32 points tested, hence a discretization approach does not find the most-likely trajectory. Surprisingly, though, the GRDRL solver also outperforms the other solvers at the 32 center-points. Despite not training from the center-points specifically, the GRDRL solver has a better average and maximum collision reward. The only degradation in performance was in collision percentage, although the GRDRL solver still outperforms the MCTS solver.

\section{Conclusion}
This paper presents a new architecture for AST to improve the validation of autonomous vehicles. The new solver treats the simulator as a black box and generalizes across a space of initial conditions. The new architecture is able to converge to a more-likely failure scenario in fewer iterations than running the discrete architecture. Future work will demonstrate performance in a high-fidelity simulator. The advancements presented in this paper show that AST is a promising approach for validating autonomous systems.  

\bibliographystyle{IEEEtran}
\bibliography{AST2}

\end{document}

%% file: Images/ASTStruct.tex
\begin{tikzpicture}[node distance=1.5cm,
    every node/.style={fill=white, font=\sffamily, text centered}, align=center]
	\node (sim)             [simulator]              {Simulator $\mathcal{S}$};
    \node (solver)          [solver, above of = sim, xshift = 0.8cm]              {Solver};
    \node (reward)          [reward, above of = sim, xshift = 5cm]              {Reward\\Function};  
  \draw[->]					(solver.west) -| node[text width=1cm, xshift = 0mm, yshift = 5mm, text centered, align=center] {Environment\\Actions} ($ (sim.north) - (15mm, 0) $);
  \draw[->] 	($(sim.east) + (0mm,2mm)$) -| node[text width=1.6cm, xshift = -17mm, yshift = 4mm] {Likelihood} ($ (reward.south) + (-2mm, 0mm) $);
  \draw[->] 	($(sim.east) + (0mm,-2mm)$) -| node[text width=1cm, xshift = -20mm, yshift = -4mm] {Event} ($ (reward.south) + (2mm, 0mm) $);
  \draw[->]		(reward.west) -- ++(0mm,0) -- node[text width=1cm, xshift = 0mm, yshift = 3mm] {Reward}(solver.east);
\end{tikzpicture}

%% file: Images/arch.tex
\begin{tikzpicture}[node distance=1.5cm,
    every node/.style={fill=white, font=\sffamily, text centered}, align=center]
    \tikzcuboid{%
    shiftx=-11cm,%
    shifty=-1cm,%
    shadeopacity=0.00,%
    scale=1.00,%
    }   
    \node (label)           [xshift = -10.55cm, yshift = -0.55cm, font=\fontsize{60}{0}\selectfont] {$\mathbf{I}$};
    \node (sim1)             [simulator, yshift = 2cm]              {Simulator $\mathcal{S}$};
    \node (solver1)          [solver, left of = sim1, xshift = -3cm]              {$\text{Solver}^{(1)}$};
    \node (sim2)             [simulator]              {Simulator $\mathcal{S}$};
    \node (solver2)          [solver, left of = sim2, xshift = -3cm]              {$\text{Solver}^{(2)}$};
    \node (sim3)             [simulator, yshift = -2cm]              {Simulator $\mathcal{S}$};
    \node (solver3)          [solver, left of = sim3, xshift = -3cm]              {$\text{Solver}^{(3)}$};
    \coordinate[left of = solver1, xshift = -2.15cm, yshift = -0.85cm] (s1)          ;
    \coordinate[left of = solver2, xshift = -2.5cm, yshift = -0.1cm] (s2)          ;
    \coordinate[left of = solver3, xshift = -3.5cm, yshift = 0.55cm] (s3)          ;
    \coordinate[left of = solver1] (c1)          ;
    \coordinate[left of = solver2] (c2)          ;
    \coordinate[left of = solver3] (c3)          ;
    \draw[->]					(s1) --node[font=\fontsize{15},text width=5mm, xshift = -2mm, yshift = 5mm, text centered, align=center, fill=none] {$s_0^{(1)}$} (c1)-- (solver1.west);
    \draw[->]					(s2) --node[font=\fontsize{15},text width=5mm, xshift = 2mm, yshift = -5mm, text centered, align=center, fill=none] {$s_0^{(2)}$} (c2)-- (solver2.west);
    \draw[->]					(s3) --node[font=\fontsize{15},text width=5mm, xshift = 2mm, yshift = -7mm, text centered, align=center, fill=none] {$s_0^{(3)}$} (c3)-- (solver3.west);
    \draw[->]					(solver1.east) --node[font=\fontsize{15},text width=5mm, xshift = 0mm, yshift = -5mm, text centered, align=center] {$a_t$} (sim1.west);
    \draw[->]					(solver2.east) --node[font=\fontsize{15},text width=5mm, xshift = 0mm, yshift = -5mm, text centered, align=center] {$a_t$} (sim2.west);
    \draw[->]					(solver3.east) --node[font=\fontsize{15},text width=5mm, xshift = 0mm, yshift = -5mm, text centered, align=center] {$a_t$} (sim3.west);
    \draw[->] 	(sim1.east) -- ($(sim1.east) + (5mm,0mm)$) --node[font=\fontsize{15},text width=5mm, xshift = 6mm, yshift = 0mm, text centered, align=center] {$s_t$}  ($(sim1.east) + (5mm,8mm)$) --  ($(c1) + (0mm,8mm)$) --  (c1)-- (solver1.west);
    \draw[->] 	(sim2.east) -- ($(sim2.east) + (5mm,0mm)$) --node[font=\fontsize{15},text width=5mm, xshift = 6mm, yshift = 0mm, text centered, align=center] {$s_t$}  ($(sim2.east) + (5mm,8mm)$) --  ($(c2) + (0mm,8mm)$) --  (c2)-- (solver2.west);
    \draw[->] 	(sim3.east) -- ($(sim3.east) + (5mm,0mm)$) --node[font=\fontsize{15},text width=5mm, xshift = 6mm, yshift = 0mm, text centered, align=center] {$s_t$}  ($(sim3.east) + (5mm,8mm)$) --  ($(c3) + (0mm,8mm)$) --  (c3)-- (solver3.west);
\end{tikzpicture}

%% file: Images/newArch.tex
\begin{tikzpicture}[node distance=1.5cm,
    every node/.style={fill=white, font=\sffamily, text centered}, align=center]
    \tikzcuboid{%
    shiftx=-11cm,%
    shifty=-1cm,%
    shadeopacity=1.00,%
    scale=1.00,
    front/.style={draw=black!10!white,fill=black!10!white},%
    top/.style={draw=black!10!white,fill=black!10!white},%
    right/.style={draw=black!10!white,fill=black!10!white},%
    emphedge,%
    emphstyle/.style={line width=1pt, green!12!black,densely dashed},
    }   
    \node (label)           [xshift = -10.55cm, yshift = -0.55cm,fill=black!10!white, font=\fontsize{60}{0}\selectfont] {$\mathbf{I}$};
    \node (sim2)             [simulator]              {Simulator $\mathcal{S}$};
    \node (solver2)          [solver, left of = sim2, xshift = -3cm]              {Solver};
    \coordinate[left of = solver2, xshift = -2.15cm, yshift = 1.15cm] (s1)          ;
    \coordinate[left of = solver2, xshift = -2.5cm, yshift = -0.1cm] (s2)          ;
    \coordinate[left of = solver2, xshift = -3.5cm, yshift = -1.45cm] (s3)          ;
    \coordinate[left of = solver2] (c2)          ;
    \draw[->]					(s1) --node[font=\fontsize{15},text width=5mm, xshift = 1mm, yshift = 6mm, text centered, align=center,  fill=none] {$s_0^{(1)}$} (c2)-- (solver2.west);
    \draw[->]					(s2) --node[font=\fontsize{15},text width=5mm, xshift = -13.5mm, yshift = 5mm, text centered, align=center,  fill=none] {$s_0^{(2)}$} (c2)-- (solver2.west);
    \draw[->]					(s3) --node[font=\fontsize{15},text width=5mm, xshift = 6mm, yshift = -4mm, text centered, align=center,  fill=none] {$s_0^{(3)}$} (c2)-- (solver2.west);
    \draw[->]					(solver2.east) --node[font=\fontsize{15},text width=5mm, xshift = 0mm, yshift = -5mm, text centered, align=center] {$a_t$} (sim2.west);
    \draw[->]                   ($(solver2.east) + (10mm,0mm)$) -- ($(solver2.east) + (10mm,8mm)$) -- ($(c2) + (0mm,8mm)$) -- (c2) -- (solver2.west) ;
    \draw [->] (solver2.south) to[in=230,out=310,loop,min distance=15mm] node[font=\fontsize{15},text width=5mm, xshift = 0mm, yshift = -5mm, text centered, align=center] {$h_t$} (solver2.south);
\end{tikzpicture}

%% file: Images/NNArch.tex
\begin{tikzpicture}[node distance=1.5cm,
    every node/.style={fill=white, font=\sffamily, text centered}, align=center]
	\node (net)             [network]              {LSTM};
    \node (input)          [io, below of = net, yshift=-0.5cm]              {$x_t$};
    \node (output)          [io, above of = net,  yshift=0.5cm]              {$y_t=[\mu_{t+1}, \Sigma_{t+1}]$};  
    \node(h_next)			[state, right of = net, xshift = 2cm] 					{$h_{t+1}$};
    \node(h_state)			[state, left of = net, xshift = -2cm] 					{$h_{t}$};
  \draw[->] 					(h_state.east) --(net.west);
  \draw[->] 					(input.north) --(net.south);
  \draw[->] 					(net.north) --(output.south);
  \draw[->] 					(net.east) --(h_next.west);
\end{tikzpicture}

%% file: Images/table_variables.tex
\npdecimalsign{.}
\nprounddigits{1}
\sisetup{round-mode=places,round-precision=2}
\begin{tabular}{@{}lrr@{}}
\toprule
      Variable & Min & Max \\ \midrule
{$\vect{s}_{0, ped, x}$}     & {$\SI{-1}{\meter}$}  &  {$\SI{1}{\meter}$}   \\
{$\vect{s}_{0, ped, y}$}      & {$\SI{-6}{\meter} $}  & {$\SI{-2}{\meter}$}  \\
{$s_{0, car}$}       & {$\SI{-43.75}{\meter}$} & {$\SI{-26.25}{\meter}$} \\
{$v_{0, ped}$}      & {$\SI{0}{\meter\per\second} $}   & {$\SI{2}{\meter\per\second}$}   \\
{$v_{0, car}$}      & {$\SI{8.34}{\meter\per\second}$}   & {$\SI{13.96}{\meter\per\second}$}                               \\ \bottomrule
\end{tabular}
\npnoround

%% file: Images/test.tex
\begin{tikzpicture}

\definecolor{color0}{rgb}{0.001462,0.000466,0.013866}
\definecolor{color1}{rgb}{0.987622,0.64532,0.039886}
\definecolor{color2}{rgb}{0.578304,0.148039,0.404411}

\begin{axis}[
font=\normalsize,
align =center,
legend cell align={left},
legend entries={{Generalized},{Discrete (Conservative)},{Discrete (Optimistic)}},
legend style={nodes={scale=0.75, transform shape}, at={(0.97,0.03)}, anchor=south east, draw=white!80.0!black},
tick align=outside,
tick pos=left,
x grid style={white!69.01960784313725!black},
xlabel={Iterations},
xmin=-25, xmax=525,
y grid style={white!69.01960784313725!black},
ylabel style={align=center}, ylabel={Mahalanobis Distance},
ymin=-200, ymax=-95
]
\addlegendimage{no markers, color0}
\addlegendimage{no markers, color1}
\addlegendimage{no markers, color2, dotted}
\addplot [thick, color0]
table [row sep=\\]{%
0	-197.177450081895 \\
1	-183.28078991658 \\
2	-149.241921054272 \\
3	-149.241921054272 \\
4	-139.93238564198 \\
5	-139.93238564198 \\
6	-139.93238564198 \\
7	-139.93238564198 \\
8	-139.93238564198 \\
9	-139.93238564198 \\
10	-139.93238564198 \\
11	-139.93238564198 \\
12	-139.93238564198 \\
13	-139.93238564198 \\
14	-139.93238564198 \\
15	-133.531327520145 \\
16	-131.354443871936 \\
17	-129.796365958594 \\
18	-129.796365958594 \\
19	-116.391383997619 \\
20	-116.391383997619 \\
21	-116.391383997619 \\
22	-116.391383997619 \\
23	-116.391383997619 \\
24	-111.229905317446 \\
25	-111.229905317446 \\
26	-111.229905317446 \\
27	-111.229905317446 \\
28	-111.229905317446 \\
29	-111.229905317446 \\
30	-111.229905317446 \\
31	-111.229905317446 \\
32	-111.229905317446 \\
33	-108.561609653249 \\
34	-108.561609653249 \\
35	-108.561609653249 \\
36	-108.561609653249 \\
37	-108.561609653249 \\
38	-108.561609653249 \\
39	-108.561609653249 \\
40	-108.561609653249 \\
41	-108.561609653249 \\
42	-108.561609653249 \\
43	-108.561609653249 \\
44	-108.561609653249 \\
45	-108.561609653249 \\
46	-108.561609653249 \\
47	-108.561609653249 \\
48	-108.561609653249 \\
49	-108.561609653249 \\
50	-108.561609653249 \\
51	-108.561609653249 \\
52	-108.561609653249 \\
53	-108.561609653249 \\
54	-105.173457790902 \\
55	-105.173457790902 \\
56	-105.173457790902 \\
57	-105.173457790902 \\
58	-105.173457790902 \\
59	-105.173457790902 \\
60	-105.173457790902 \\
61	-105.173457790902 \\
62	-105.173457790902 \\
63	-105.173457790902 \\
64	-105.173457790902 \\
65	-105.173457790902 \\
66	-105.173457790902 \\
67	-105.173457790902 \\
68	-105.173457790902 \\
69	-105.173457790902 \\
70	-105.173457790902 \\
71	-105.173457790902 \\
72	-105.173457790902 \\
73	-105.173457790902 \\
74	-105.173457790902 \\
75	-105.173457790902 \\
76	-105.173457790902 \\
77	-105.173457790902 \\
78	-105.173457790902 \\
79	-105.173457790902 \\
80	-105.173457790902 \\
81	-105.173457790902 \\
82	-105.173457790902 \\
83	-100.9901809119 \\
84	-100.9901809119 \\
85	-100.9901809119 \\
86	-100.9901809119 \\
87	-100.9901809119 \\
88	-100.9901809119 \\
89	-100.9901809119 \\
90	-100.9901809119 \\
91	-100.9901809119 \\
92	-100.9901809119 \\
93	-100.9901809119 \\
94	-100.9901809119 \\
95	-100.9901809119 \\
96	-100.9901809119 \\
97	-100.9901809119 \\
98	-100.9901809119 \\
99	-100.9901809119 \\
100	-100.9901809119 \\
101	-100.9901809119 \\
102	-100.9901809119 \\
103	-100.9901809119 \\
104	-100.9901809119 \\
105	-100.9901809119 \\
106	-100.9901809119 \\
107	-100.9901809119 \\
108	-100.9901809119 \\
109	-100.9901809119 \\
110	-100.9901809119 \\
111	-100.9901809119 \\
112	-100.9901809119 \\
113	-100.9901809119 \\
114	-100.9901809119 \\
115	-100.9901809119 \\
116	-100.9901809119 \\
117	-100.9901809119 \\
118	-100.9901809119 \\
119	-100.9901809119 \\
120	-100.9901809119 \\
121	-100.9901809119 \\
122	-100.9901809119 \\
123	-100.9901809119 \\
124	-100.9901809119 \\
125	-100.9901809119 \\
126	-100.9901809119 \\
127	-100.9901809119 \\
128	-100.9901809119 \\
129	-100.9901809119 \\
130	-100.9901809119 \\
131	-100.9901809119 \\
132	-100.9901809119 \\
133	-100.9901809119 \\
134	-100.9901809119 \\
135	-100.9901809119 \\
136	-100.9901809119 \\
137	-100.9901809119 \\
138	-100.9901809119 \\
139	-100.9901809119 \\
140	-100.9901809119 \\
141	-100.9901809119 \\
142	-100.9901809119 \\
143	-100.9901809119 \\
144	-100.9901809119 \\
145	-100.9901809119 \\
146	-100.9901809119 \\
147	-100.9901809119 \\
148	-100.9901809119 \\
149	-100.9901809119 \\
150	-100.9901809119 \\
151	-100.9901809119 \\
152	-100.9901809119 \\
153	-100.9901809119 \\
154	-100.9901809119 \\
155	-100.9901809119 \\
156	-100.9901809119 \\
157	-100.9901809119 \\
158	-100.9901809119 \\
159	-100.9901809119 \\
160	-100.9901809119 \\
161	-100.9901809119 \\
162	-100.9901809119 \\
163	-100.9901809119 \\
164	-100.9901809119 \\
165	-100.9901809119 \\
166	-100.9901809119 \\
167	-100.9901809119 \\
168	-100.9901809119 \\
169	-100.9901809119 \\
170	-100.9901809119 \\
171	-100.9901809119 \\
172	-100.9901809119 \\
173	-100.9901809119 \\
174	-100.9901809119 \\
175	-100.9901809119 \\
176	-100.9901809119 \\
177	-100.9901809119 \\
178	-100.9901809119 \\
179	-100.9901809119 \\
180	-100.9901809119 \\
181	-100.9901809119 \\
182	-100.9901809119 \\
183	-100.9901809119 \\
184	-100.9901809119 \\
185	-100.9901809119 \\
186	-100.9901809119 \\
187	-100.9901809119 \\
188	-100.9901809119 \\
189	-100.9901809119 \\
190	-100.9901809119 \\
191	-100.9901809119 \\
192	-100.9901809119 \\
193	-100.9901809119 \\
194	-100.9901809119 \\
195	-100.9901809119 \\
196	-100.9901809119 \\
197	-100.9901809119 \\
198	-100.9901809119 \\
199	-100.9901809119 \\
200	-100.9901809119 \\
201	-100.9901809119 \\
202	-100.9901809119 \\
203	-100.9901809119 \\
204	-100.9901809119 \\
205	-100.9901809119 \\
206	-100.9901809119 \\
207	-100.9901809119 \\
208	-100.9901809119 \\
209	-100.9901809119 \\
210	-100.9901809119 \\
211	-100.9901809119 \\
212	-100.9901809119 \\
213	-100.9901809119 \\
214	-100.9901809119 \\
215	-100.9901809119 \\
216	-100.9901809119 \\
217	-100.9901809119 \\
218	-100.9901809119 \\
219	-100.9901809119 \\
220	-100.9901809119 \\
221	-100.9901809119 \\
222	-100.9901809119 \\
223	-100.9901809119 \\
224	-100.9901809119 \\
225	-100.9901809119 \\
226	-100.9901809119 \\
227	-100.9901809119 \\
228	-100.9901809119 \\
229	-100.9901809119 \\
230	-100.9901809119 \\
231	-100.9901809119 \\
232	-100.9901809119 \\
233	-100.9901809119 \\
234	-100.9901809119 \\
235	-100.9901809119 \\
236	-100.9901809119 \\
237	-100.9901809119 \\
238	-100.9901809119 \\
239	-100.9901809119 \\
240	-100.9901809119 \\
241	-100.9901809119 \\
242	-100.9901809119 \\
243	-100.9901809119 \\
244	-100.9901809119 \\
245	-100.9901809119 \\
246	-100.9901809119 \\
247	-100.9901809119 \\
248	-100.9901809119 \\
249	-100.9901809119 \\
250	-100.9901809119 \\
251	-100.9901809119 \\
252	-100.9901809119 \\
253	-100.9901809119 \\
254	-100.9901809119 \\
255	-100.9901809119 \\
256	-100.9901809119 \\
257	-100.9901809119 \\
258	-100.9901809119 \\
259	-100.9901809119 \\
260	-100.9901809119 \\
261	-100.9901809119 \\
262	-100.9901809119 \\
263	-100.9901809119 \\
264	-100.9901809119 \\
265	-100.9901809119 \\
266	-100.9901809119 \\
267	-100.9901809119 \\
268	-100.9901809119 \\
269	-100.9901809119 \\
270	-100.9901809119 \\
271	-100.9901809119 \\
272	-100.9901809119 \\
273	-100.9901809119 \\
274	-100.9901809119 \\
275	-100.9901809119 \\
276	-100.9901809119 \\
277	-100.9901809119 \\
278	-100.9901809119 \\
279	-100.9901809119 \\
280	-100.9901809119 \\
281	-100.9901809119 \\
282	-100.9901809119 \\
283	-100.9901809119 \\
284	-100.9901809119 \\
285	-100.9901809119 \\
286	-100.9901809119 \\
287	-100.9901809119 \\
288	-100.9901809119 \\
289	-100.9901809119 \\
290	-100.9901809119 \\
291	-100.9901809119 \\
292	-100.9901809119 \\
293	-100.9901809119 \\
294	-100.9901809119 \\
295	-100.9901809119 \\
296	-100.9901809119 \\
297	-100.9901809119 \\
298	-100.9901809119 \\
299	-100.9901809119 \\
300	-100.9901809119 \\
301	-100.9901809119 \\
302	-100.9901809119 \\
303	-100.9901809119 \\
304	-100.9901809119 \\
305	-100.9901809119 \\
306	-100.9901809119 \\
307	-100.9901809119 \\
308	-100.9901809119 \\
309	-100.9901809119 \\
310	-100.9901809119 \\
311	-100.9901809119 \\
312	-100.9901809119 \\
313	-100.9901809119 \\
314	-100.9901809119 \\
315	-100.9901809119 \\
316	-100.9901809119 \\
317	-100.9901809119 \\
318	-100.9901809119 \\
319	-100.9901809119 \\
320	-100.9901809119 \\
321	-100.9901809119 \\
322	-100.9901809119 \\
323	-100.9901809119 \\
324	-100.9901809119 \\
325	-100.9901809119 \\
326	-100.9901809119 \\
327	-100.9901809119 \\
328	-100.9901809119 \\
329	-100.9901809119 \\
330	-100.9901809119 \\
331	-100.9901809119 \\
332	-100.9901809119 \\
333	-100.9901809119 \\
334	-100.9901809119 \\
335	-100.9901809119 \\
336	-100.9901809119 \\
337	-100.9901809119 \\
338	-100.9901809119 \\
339	-100.9901809119 \\
340	-100.9901809119 \\
341	-100.9901809119 \\
342	-100.9901809119 \\
343	-100.9901809119 \\
344	-100.9901809119 \\
345	-100.9901809119 \\
346	-100.9901809119 \\
347	-100.9901809119 \\
348	-100.9901809119 \\
349	-100.9901809119 \\
350	-100.9901809119 \\
351	-100.9901809119 \\
352	-100.9901809119 \\
353	-100.9901809119 \\
354	-100.9901809119 \\
355	-100.9901809119 \\
356	-100.9901809119 \\
357	-100.9901809119 \\
358	-100.9901809119 \\
359	-100.9901809119 \\
360	-100.9901809119 \\
361	-100.9901809119 \\
362	-100.9901809119 \\
363	-100.9901809119 \\
364	-100.9901809119 \\
365	-100.9901809119 \\
366	-100.9901809119 \\
367	-100.9901809119 \\
368	-100.9901809119 \\
369	-100.9901809119 \\
370	-100.9901809119 \\
371	-100.9901809119 \\
372	-100.9901809119 \\
373	-100.9901809119 \\
374	-100.9901809119 \\
375	-100.9901809119 \\
376	-100.9901809119 \\
377	-100.9901809119 \\
378	-100.9901809119 \\
379	-100.9901809119 \\
380	-100.9901809119 \\
381	-100.9901809119 \\
382	-100.9901809119 \\
383	-100.9901809119 \\
384	-100.9901809119 \\
385	-100.9901809119 \\
386	-100.9901809119 \\
387	-100.9901809119 \\
388	-100.9901809119 \\
389	-100.9901809119 \\
390	-100.9901809119 \\
391	-100.9901809119 \\
392	-100.9901809119 \\
393	-100.9901809119 \\
394	-100.9901809119 \\
395	-100.9901809119 \\
396	-100.9901809119 \\
397	-100.9901809119 \\
398	-100.9901809119 \\
399	-100.9901809119 \\
400	-100.9901809119 \\
401	-100.9901809119 \\
402	-100.9901809119 \\
403	-100.9901809119 \\
404	-100.9901809119 \\
405	-100.9901809119 \\
406	-100.9901809119 \\
407	-100.9901809119 \\
408	-100.9901809119 \\
409	-100.9901809119 \\
410	-100.9901809119 \\
411	-100.9901809119 \\
412	-100.9901809119 \\
413	-100.9901809119 \\
414	-100.9901809119 \\
415	-100.9901809119 \\
416	-100.9901809119 \\
417	-100.9901809119 \\
418	-100.9901809119 \\
419	-100.9901809119 \\
420	-100.9901809119 \\
421	-100.9901809119 \\
422	-100.9901809119 \\
423	-100.9901809119 \\
424	-100.9901809119 \\
425	-100.9901809119 \\
426	-100.9901809119 \\
427	-100.9901809119 \\
428	-100.9901809119 \\
429	-100.9901809119 \\
430	-100.9901809119 \\
431	-100.9901809119 \\
432	-100.9901809119 \\
433	-100.9901809119 \\
434	-100.9901809119 \\
435	-100.9901809119 \\
436	-100.9901809119 \\
437	-100.9901809119 \\
438	-100.9901809119 \\
439	-100.9901809119 \\
440	-100.9901809119 \\
441	-100.9901809119 \\
442	-100.9901809119 \\
443	-100.9901809119 \\
444	-100.9901809119 \\
445	-100.9901809119 \\
446	-100.9901809119 \\
447	-100.9901809119 \\
448	-100.9901809119 \\
449	-100.9901809119 \\
450	-100.9901809119 \\
451	-100.9901809119 \\
452	-100.9901809119 \\
453	-100.9901809119 \\
454	-100.9901809119 \\
455	-100.9901809119 \\
456	-100.9901809119 \\
457	-100.9901809119 \\
458	-100.9901809119 \\
459	-100.9901809119 \\
460	-100.9901809119 \\
461	-100.9901809119 \\
462	-100.9901809119 \\
463	-100.9901809119 \\
464	-100.9901809119 \\
465	-100.9901809119 \\
466	-100.9901809119 \\
467	-100.9901809119 \\
468	-100.9901809119 \\
469	-100.9901809119 \\
470	-100.9901809119 \\
471	-100.9901809119 \\
472	-100.9901809119 \\
473	-100.9901809119 \\
474	-100.9901809119 \\
475	-100.9901809119 \\
476	-100.9901809119 \\
477	-100.9901809119 \\
478	-100.9901809119 \\
479	-100.9901809119 \\
480	-100.9901809119 \\
481	-100.9901809119 \\
482	-100.9901809119 \\
483	-100.9901809119 \\
484	-100.9901809119 \\
485	-100.9901809119 \\
486	-100.9901809119 \\
487	-100.9901809119 \\
488	-100.9901809119 \\
489	-100.9901809119 \\
490	-100.9901809119 \\
491	-100.9901809119 \\
492	-100.9901809119 \\
493	-100.9901809119 \\
494	-100.9901809119 \\
495	-100.9901809119 \\
496	-100.9901809119 \\
497	-100.9901809119 \\
498	-100.9901809119 \\
499	-100.9901809119 \\
500	-100.9901809119 \\
};
\addplot [thick, color1]
table [row sep=\\]{%
0	-284.561767456417 \\
1	-177.055822903157 \\
2	-177.055822903157 \\
3	-177.055822903157 \\
4	-177.055822903157 \\
5	-177.055822903157 \\
6	-177.055822903157 \\
7	-177.055822903157 \\
8	-177.055822903157 \\
9	-177.055822903157 \\
10	-177.055822903157 \\
11	-177.055822903157 \\
12	-177.055822903157 \\
13	-177.055822903157 \\
14	-157.490690332793 \\
15	-157.490690332793 \\
16	-157.490690332793 \\
17	-157.490690332793 \\
18	-157.490690332793 \\
19	-157.490690332793 \\
20	-157.490690332793 \\
21	-157.490690332793 \\
22	-157.490690332793 \\
23	-157.490690332793 \\
24	-157.490690332793 \\
25	-157.490690332793 \\
26	-157.490690332793 \\
27	-157.490690332793 \\
28	-157.490690332793 \\
29	-157.490690332793 \\
30	-157.490690332793 \\
31	-157.490690332793 \\
32	-157.490690332793 \\
33	-157.490690332793 \\
34	-157.490690332793 \\
35	-157.490690332793 \\
36	-157.490690332793 \\
37	-157.490690332793 \\
38	-157.490690332793 \\
39	-157.490690332793 \\
40	-157.490690332793 \\
41	-157.490690332793 \\
42	-157.490690332793 \\
43	-157.490690332793 \\
44	-157.490690332793 \\
45	-157.490690332793 \\
46	-148.299776554396 \\
47	-148.299776554396 \\
48	-148.299776554396 \\
49	-148.299776554396 \\
50	-148.299776554396 \\
51	-148.299776554396 \\
52	-148.299776554396 \\
53	-148.299776554396 \\
54	-148.299776554396 \\
55	-148.299776554396 \\
56	-148.299776554396 \\
57	-148.299776554396 \\
58	-148.299776554396 \\
59	-148.299776554396 \\
60	-148.299776554396 \\
61	-148.299776554396 \\
62	-148.299776554396 \\
63	-148.299776554396 \\
64	-148.299776554396 \\
65	-148.299776554396 \\
66	-148.299776554396 \\
67	-148.299776554396 \\
68	-148.299776554396 \\
69	-148.299776554396 \\
70	-146.730359996363 \\
71	-146.730359996363 \\
72	-146.730359996363 \\
73	-146.730359996363 \\
74	-146.730359996363 \\
75	-146.730359996363 \\
76	-146.730359996363 \\
77	-146.730359996363 \\
78	-138.74229869082 \\
79	-138.74229869082 \\
80	-138.74229869082 \\
81	-138.74229869082 \\
82	-138.74229869082 \\
83	-138.74229869082 \\
84	-138.74229869082 \\
85	-138.74229869082 \\
86	-138.74229869082 \\
87	-138.74229869082 \\
88	-138.74229869082 \\
89	-138.74229869082 \\
90	-138.74229869082 \\
91	-138.74229869082 \\
92	-138.74229869082 \\
93	-138.74229869082 \\
94	-138.74229869082 \\
95	-138.74229869082 \\
96	-138.74229869082 \\
97	-138.74229869082 \\
98	-138.74229869082 \\
99	-138.74229869082 \\
100	-138.74229869082 \\
101	-138.74229869082 \\
102	-138.74229869082 \\
103	-138.74229869082 \\
104	-138.74229869082 \\
105	-138.74229869082 \\
106	-138.74229869082 \\
107	-138.74229869082 \\
108	-138.74229869082 \\
109	-138.74229869082 \\
110	-138.74229869082 \\
111	-138.74229869082 \\
112	-138.74229869082 \\
113	-138.74229869082 \\
114	-138.74229869082 \\
115	-138.74229869082 \\
116	-138.74229869082 \\
117	-138.74229869082 \\
118	-138.74229869082 \\
119	-138.74229869082 \\
120	-138.74229869082 \\
121	-138.74229869082 \\
122	-138.74229869082 \\
123	-138.74229869082 \\
124	-138.74229869082 \\
125	-138.74229869082 \\
126	-138.74229869082 \\
127	-138.74229869082 \\
128	-138.74229869082 \\
129	-138.74229869082 \\
130	-138.74229869082 \\
131	-138.74229869082 \\
132	-138.74229869082 \\
133	-138.74229869082 \\
134	-138.74229869082 \\
135	-138.74229869082 \\
136	-138.74229869082 \\
137	-138.74229869082 \\
138	-138.74229869082 \\
139	-138.74229869082 \\
140	-138.74229869082 \\
141	-138.74229869082 \\
142	-138.74229869082 \\
143	-138.74229869082 \\
144	-138.74229869082 \\
145	-138.74229869082 \\
146	-138.74229869082 \\
147	-138.74229869082 \\
148	-138.74229869082 \\
149	-138.74229869082 \\
150	-138.74229869082 \\
151	-138.74229869082 \\
152	-138.74229869082 \\
153	-138.74229869082 \\
154	-138.74229869082 \\
155	-138.74229869082 \\
156	-138.74229869082 \\
157	-138.74229869082 \\
158	-138.74229869082 \\
159	-138.74229869082 \\
160	-138.74229869082 \\
161	-138.74229869082 \\
162	-138.74229869082 \\
163	-138.74229869082 \\
164	-138.74229869082 \\
165	-138.74229869082 \\
166	-138.74229869082 \\
167	-138.74229869082 \\
168	-138.74229869082 \\
169	-138.74229869082 \\
170	-138.74229869082 \\
171	-138.74229869082 \\
172	-138.74229869082 \\
173	-138.74229869082 \\
174	-138.74229869082 \\
175	-138.74229869082 \\
176	-138.74229869082 \\
177	-138.74229869082 \\
178	-138.74229869082 \\
179	-138.74229869082 \\
180	-138.74229869082 \\
181	-138.74229869082 \\
182	-138.74229869082 \\
183	-138.74229869082 \\
184	-138.74229869082 \\
185	-138.74229869082 \\
186	-138.74229869082 \\
187	-138.74229869082 \\
188	-138.74229869082 \\
189	-138.74229869082 \\
190	-138.74229869082 \\
191	-138.74229869082 \\
192	-138.74229869082 \\
193	-138.74229869082 \\
194	-138.74229869082 \\
195	-138.74229869082 \\
196	-138.74229869082 \\
197	-138.74229869082 \\
198	-138.74229869082 \\
199	-138.74229869082 \\
200	-138.74229869082 \\
201	-133.375913725193 \\
202	-133.375913725193 \\
203	-133.375913725193 \\
204	-133.375913725193 \\
205	-133.375913725193 \\
206	-133.375913725193 \\
207	-133.375913725193 \\
208	-133.375913725193 \\
209	-133.375913725193 \\
210	-133.375913725193 \\
211	-133.375913725193 \\
212	-133.375913725193 \\
213	-133.375913725193 \\
214	-133.375913725193 \\
215	-133.375913725193 \\
216	-133.375913725193 \\
217	-133.375913725193 \\
218	-133.375913725193 \\
219	-133.375913725193 \\
220	-133.375913725193 \\
221	-133.375913725193 \\
222	-133.375913725193 \\
223	-133.375913725193 \\
224	-133.375913725193 \\
225	-133.375913725193 \\
226	-133.375913725193 \\
227	-133.375913725193 \\
228	-133.375913725193 \\
229	-133.375913725193 \\
230	-133.375913725193 \\
231	-133.375913725193 \\
232	-133.375913725193 \\
233	-123.692119067383 \\
234	-123.692119067383 \\
235	-123.692119067383 \\
236	-123.692119067383 \\
237	-123.692119067383 \\
238	-123.692119067383 \\
239	-123.692119067383 \\
240	-123.692119067383 \\
241	-123.692119067383 \\
242	-123.692119067383 \\
243	-123.692119067383 \\
244	-123.692119067383 \\
245	-123.692119067383 \\
246	-123.692119067383 \\
247	-123.692119067383 \\
248	-123.692119067383 \\
249	-123.692119067383 \\
250	-123.692119067383 \\
251	-123.692119067383 \\
252	-123.692119067383 \\
253	-123.692119067383 \\
254	-123.692119067383 \\
255	-123.692119067383 \\
256	-123.692119067383 \\
257	-123.692119067383 \\
258	-123.692119067383 \\
259	-123.692119067383 \\
260	-123.692119067383 \\
261	-123.692119067383 \\
262	-123.692119067383 \\
263	-123.692119067383 \\
264	-123.692119067383 \\
265	-123.440679572952 \\
266	-123.440679572952 \\
267	-123.440679572952 \\
268	-123.440679572952 \\
269	-123.440679572952 \\
270	-123.440679572952 \\
271	-123.440679572952 \\
272	-123.440679572952 \\
273	-123.440679572952 \\
274	-123.440679572952 \\
275	-123.440679572952 \\
276	-123.440679572952 \\
277	-123.440679572952 \\
278	-123.440679572952 \\
279	-123.440679572952 \\
280	-123.440679572952 \\
281	-123.440679572952 \\
282	-123.440679572952 \\
283	-123.440679572952 \\
284	-123.440679572952 \\
285	-123.440679572952 \\
286	-123.440679572952 \\
287	-123.440679572952 \\
288	-123.440679572952 \\
289	-123.440679572952 \\
290	-123.440679572952 \\
291	-123.440679572952 \\
292	-123.440679572952 \\
293	-123.440679572952 \\
294	-123.440679572952 \\
295	-123.440679572952 \\
296	-123.440679572952 \\
297	-120.878850198848 \\
298	-120.878850198848 \\
299	-120.878850198848 \\
300	-120.878850198848 \\
301	-120.878850198848 \\
302	-120.878850198848 \\
303	-120.878850198848 \\
304	-120.878850198848 \\
305	-120.878850198848 \\
306	-120.878850198848 \\
307	-120.878850198848 \\
308	-120.878850198848 \\
309	-120.878850198848 \\
310	-120.878850198848 \\
311	-120.878850198848 \\
312	-120.878850198848 \\
313	-120.878850198848 \\
314	-120.878850198848 \\
315	-120.878850198848 \\
316	-120.878850198848 \\
317	-120.878850198848 \\
318	-120.878850198848 \\
319	-120.878850198848 \\
320	-120.878850198848 \\
321	-120.878850198848 \\
322	-120.878850198848 \\
323	-120.878850198848 \\
324	-120.878850198848 \\
325	-120.878850198848 \\
326	-120.878850198848 \\
327	-120.878850198848 \\
328	-120.878850198848 \\
329	-118.206167232445 \\
330	-118.206167232445 \\
331	-118.206167232445 \\
332	-118.206167232445 \\
333	-118.206167232445 \\
334	-118.206167232445 \\
335	-118.206167232445 \\
336	-118.206167232445 \\
337	-118.206167232445 \\
338	-118.206167232445 \\
339	-118.206167232445 \\
340	-118.206167232445 \\
341	-118.206167232445 \\
342	-118.206167232445 \\
343	-118.206167232445 \\
344	-118.206167232445 \\
345	-118.206167232445 \\
346	-118.206167232445 \\
347	-118.206167232445 \\
348	-118.206167232445 \\
349	-118.206167232445 \\
350	-118.206167232445 \\
351	-118.206167232445 \\
352	-118.206167232445 \\
353	-118.206167232445 \\
354	-118.206167232445 \\
355	-118.206167232445 \\
356	-118.206167232445 \\
357	-118.206167232445 \\
358	-118.206167232445 \\
359	-118.206167232445 \\
360	-118.206167232445 \\
361	-116.376780946552 \\
362	-116.376780946552 \\
363	-116.376780946552 \\
364	-116.376780946552 \\
365	-116.376780946552 \\
366	-116.376780946552 \\
367	-116.376780946552 \\
368	-116.376780946552 \\
369	-116.376780946552 \\
370	-116.376780946552 \\
371	-116.376780946552 \\
372	-116.376780946552 \\
373	-116.376780946552 \\
374	-116.376780946552 \\
375	-116.376780946552 \\
376	-116.376780946552 \\
377	-116.376780946552 \\
378	-116.376780946552 \\
379	-116.376780946552 \\
380	-116.376780946552 \\
381	-116.376780946552 \\
382	-116.376780946552 \\
383	-116.376780946552 \\
384	-116.376780946552 \\
385	-116.376780946552 \\
386	-116.376780946552 \\
387	-116.376780946552 \\
388	-116.376780946552 \\
389	-116.376780946552 \\
390	-116.376780946552 \\
391	-116.376780946552 \\
392	-116.376780946552 \\
393	-116.376780946552 \\
394	-116.376780946552 \\
395	-116.376780946552 \\
396	-116.376780946552 \\
397	-116.376780946552 \\
398	-116.376780946552 \\
399	-116.376780946552 \\
400	-116.376780946552 \\
401	-116.376780946552 \\
402	-116.376780946552 \\
403	-116.376780946552 \\
404	-116.376780946552 \\
405	-116.376780946552 \\
406	-116.376780946552 \\
407	-116.376780946552 \\
408	-116.376780946552 \\
409	-116.376780946552 \\
410	-116.376780946552 \\
411	-116.376780946552 \\
412	-116.376780946552 \\
413	-116.376780946552 \\
414	-116.376780946552 \\
415	-116.376780946552 \\
416	-116.376780946552 \\
417	-116.376780946552 \\
418	-116.376780946552 \\
419	-116.376780946552 \\
420	-116.376780946552 \\
421	-116.376780946552 \\
422	-116.376780946552 \\
423	-116.376780946552 \\
424	-116.376780946552 \\
425	-114.582657425991 \\
426	-114.582657425991 \\
427	-114.582657425991 \\
428	-114.582657425991 \\
429	-114.582657425991 \\
430	-114.582657425991 \\
431	-114.582657425991 \\
432	-114.582657425991 \\
433	-114.582657425991 \\
434	-114.582657425991 \\
435	-114.582657425991 \\
436	-114.582657425991 \\
437	-114.582657425991 \\
438	-114.582657425991 \\
439	-114.582657425991 \\
440	-114.582657425991 \\
441	-114.582657425991 \\
442	-114.582657425991 \\
443	-114.582657425991 \\
444	-114.582657425991 \\
445	-114.582657425991 \\
446	-114.582657425991 \\
447	-114.582657425991 \\
448	-114.582657425991 \\
449	-114.582657425991 \\
450	-114.582657425991 \\
451	-114.582657425991 \\
452	-114.582657425991 \\
453	-114.582657425991 \\
454	-114.582657425991 \\
455	-114.582657425991 \\
456	-114.582657425991 \\
457	-114.18531858492 \\
458	-114.18531858492 \\
459	-114.18531858492 \\
460	-114.18531858492 \\
461	-114.18531858492 \\
462	-114.18531858492 \\
463	-114.18531858492 \\
464	-114.18531858492 \\
465	-114.18531858492 \\
466	-114.18531858492 \\
467	-114.18531858492 \\
468	-114.18531858492 \\
469	-114.18531858492 \\
470	-114.18531858492 \\
471	-114.18531858492 \\
472	-114.18531858492 \\
473	-114.18531858492 \\
474	-114.18531858492 \\
475	-114.18531858492 \\
476	-114.18531858492 \\
477	-114.18531858492 \\
478	-114.18531858492 \\
479	-114.18531858492 \\
480	-114.18531858492 \\
481	-114.18531858492 \\
482	-114.18531858492 \\
483	-114.18531858492 \\
484	-114.18531858492 \\
485	-114.18531858492 \\
486	-114.18531858492 \\
487	-114.18531858492 \\
488	-114.18531858492 \\
489	-114.18531858492 \\
490	-114.18531858492 \\
491	-114.18531858492 \\
492	-114.18531858492 \\
493	-114.18531858492 \\
494	-114.18531858492 \\
495	-114.18531858492 \\
496	-114.18531858492 \\
497	-114.18531858492 \\
498	-114.18531858492 \\
499	-114.18531858492 \\
500	-114.18531858492 \\
};
\addplot [line width=1.0pt, color2, dotted]
table [row sep=\\]{%
0	-157.490690332793 \\
1	-157.490690332793 \\
2	-157.490690332793 \\
3	-157.490690332793 \\
4	-157.490690332793 \\
5	-157.490690332793 \\
6	-157.490690332793 \\
7	-157.490690332793 \\
8	-157.490690332793 \\
9	-157.490690332793 \\
10	-157.490690332793 \\
11	-157.490690332793 \\
12	-157.490690332793 \\
13	-157.490690332793 \\
14	-157.490690332793 \\
15	-157.490690332793 \\
16	-157.490690332793 \\
17	-157.490690332793 \\
18	-157.490690332793 \\
19	-157.490690332793 \\
20	-157.490690332793 \\
21	-157.490690332793 \\
22	-157.490690332793 \\
23	-157.490690332793 \\
24	-157.490690332793 \\
25	-157.490690332793 \\
26	-157.490690332793 \\
27	-157.490690332793 \\
28	-157.490690332793 \\
29	-157.490690332793 \\
30	-157.490690332793 \\
31	-157.490690332793 \\
32	-148.299776554396 \\
33	-148.299776554396 \\
34	-148.299776554396 \\
35	-148.299776554396 \\
36	-148.299776554396 \\
37	-148.299776554396 \\
38	-148.299776554396 \\
39	-148.299776554396 \\
40	-148.299776554396 \\
41	-148.299776554396 \\
42	-148.299776554396 \\
43	-148.299776554396 \\
44	-148.299776554396 \\
45	-148.299776554396 \\
46	-148.299776554396 \\
47	-148.299776554396 \\
48	-148.299776554396 \\
49	-148.299776554396 \\
50	-148.299776554396 \\
51	-148.299776554396 \\
52	-148.299776554396 \\
53	-148.299776554396 \\
54	-148.299776554396 \\
55	-148.299776554396 \\
56	-148.299776554396 \\
57	-148.299776554396 \\
58	-148.299776554396 \\
59	-148.299776554396 \\
60	-148.299776554396 \\
61	-148.299776554396 \\
62	-148.299776554396 \\
63	-148.299776554396 \\
64	-138.74229869082 \\
65	-138.74229869082 \\
66	-138.74229869082 \\
67	-138.74229869082 \\
68	-138.74229869082 \\
69	-138.74229869082 \\
70	-138.74229869082 \\
71	-138.74229869082 \\
72	-138.74229869082 \\
73	-138.74229869082 \\
74	-138.74229869082 \\
75	-138.74229869082 \\
76	-138.74229869082 \\
77	-138.74229869082 \\
78	-138.74229869082 \\
79	-138.74229869082 \\
80	-138.74229869082 \\
81	-138.74229869082 \\
82	-138.74229869082 \\
83	-138.74229869082 \\
84	-138.74229869082 \\
85	-138.74229869082 \\
86	-138.74229869082 \\
87	-138.74229869082 \\
88	-138.74229869082 \\
89	-138.74229869082 \\
90	-138.74229869082 \\
91	-138.74229869082 \\
92	-138.74229869082 \\
93	-138.74229869082 \\
94	-138.74229869082 \\
95	-138.74229869082 \\
96	-138.74229869082 \\
97	-138.74229869082 \\
98	-138.74229869082 \\
99	-138.74229869082 \\
100	-138.74229869082 \\
101	-138.74229869082 \\
102	-138.74229869082 \\
103	-138.74229869082 \\
104	-138.74229869082 \\
105	-138.74229869082 \\
106	-138.74229869082 \\
107	-138.74229869082 \\
108	-138.74229869082 \\
109	-138.74229869082 \\
110	-138.74229869082 \\
111	-138.74229869082 \\
112	-138.74229869082 \\
113	-138.74229869082 \\
114	-138.74229869082 \\
115	-138.74229869082 \\
116	-138.74229869082 \\
117	-138.74229869082 \\
118	-138.74229869082 \\
119	-138.74229869082 \\
120	-138.74229869082 \\
121	-138.74229869082 \\
122	-138.74229869082 \\
123	-138.74229869082 \\
124	-138.74229869082 \\
125	-138.74229869082 \\
126	-138.74229869082 \\
127	-138.74229869082 \\
128	-138.74229869082 \\
129	-138.74229869082 \\
130	-138.74229869082 \\
131	-138.74229869082 \\
132	-138.74229869082 \\
133	-138.74229869082 \\
134	-138.74229869082 \\
135	-138.74229869082 \\
136	-138.74229869082 \\
137	-138.74229869082 \\
138	-138.74229869082 \\
139	-138.74229869082 \\
140	-138.74229869082 \\
141	-138.74229869082 \\
142	-138.74229869082 \\
143	-138.74229869082 \\
144	-138.74229869082 \\
145	-138.74229869082 \\
146	-138.74229869082 \\
147	-138.74229869082 \\
148	-138.74229869082 \\
149	-138.74229869082 \\
150	-138.74229869082 \\
151	-138.74229869082 \\
152	-138.74229869082 \\
153	-138.74229869082 \\
154	-138.74229869082 \\
155	-138.74229869082 \\
156	-138.74229869082 \\
157	-138.74229869082 \\
158	-138.74229869082 \\
159	-138.74229869082 \\
160	-138.74229869082 \\
161	-138.74229869082 \\
162	-138.74229869082 \\
163	-138.74229869082 \\
164	-138.74229869082 \\
165	-138.74229869082 \\
166	-138.74229869082 \\
167	-138.74229869082 \\
168	-138.74229869082 \\
169	-138.74229869082 \\
170	-138.74229869082 \\
171	-138.74229869082 \\
172	-138.74229869082 \\
173	-138.74229869082 \\
174	-138.74229869082 \\
175	-138.74229869082 \\
176	-138.74229869082 \\
177	-138.74229869082 \\
178	-138.74229869082 \\
179	-138.74229869082 \\
180	-138.74229869082 \\
181	-138.74229869082 \\
182	-138.74229869082 \\
183	-138.74229869082 \\
184	-138.74229869082 \\
185	-138.74229869082 \\
186	-138.74229869082 \\
187	-138.74229869082 \\
188	-138.74229869082 \\
189	-138.74229869082 \\
190	-138.74229869082 \\
191	-138.74229869082 \\
192	-133.375913725193 \\
193	-133.375913725193 \\
194	-133.375913725193 \\
195	-133.375913725193 \\
196	-133.375913725193 \\
197	-133.375913725193 \\
198	-133.375913725193 \\
199	-133.375913725193 \\
200	-133.375913725193 \\
201	-133.375913725193 \\
202	-133.375913725193 \\
203	-133.375913725193 \\
204	-133.375913725193 \\
205	-133.375913725193 \\
206	-133.375913725193 \\
207	-133.375913725193 \\
208	-133.375913725193 \\
209	-133.375913725193 \\
210	-133.375913725193 \\
211	-133.375913725193 \\
212	-133.375913725193 \\
213	-133.375913725193 \\
214	-133.375913725193 \\
215	-133.375913725193 \\
216	-133.375913725193 \\
217	-133.375913725193 \\
218	-133.375913725193 \\
219	-133.375913725193 \\
220	-133.375913725193 \\
221	-133.375913725193 \\
222	-133.375913725193 \\
223	-133.375913725193 \\
224	-123.692119067383 \\
225	-123.692119067383 \\
226	-123.692119067383 \\
227	-123.692119067383 \\
228	-123.692119067383 \\
229	-123.692119067383 \\
230	-123.692119067383 \\
231	-123.692119067383 \\
232	-123.692119067383 \\
233	-123.692119067383 \\
234	-123.692119067383 \\
235	-123.692119067383 \\
236	-123.692119067383 \\
237	-123.692119067383 \\
238	-123.692119067383 \\
239	-123.692119067383 \\
240	-123.692119067383 \\
241	-123.692119067383 \\
242	-123.692119067383 \\
243	-123.692119067383 \\
244	-123.692119067383 \\
245	-123.692119067383 \\
246	-123.692119067383 \\
247	-123.692119067383 \\
248	-123.692119067383 \\
249	-123.692119067383 \\
250	-123.692119067383 \\
251	-123.692119067383 \\
252	-123.692119067383 \\
253	-123.692119067383 \\
254	-123.692119067383 \\
255	-123.692119067383 \\
256	-123.440679572952 \\
257	-123.440679572952 \\
258	-123.440679572952 \\
259	-123.440679572952 \\
260	-123.440679572952 \\
261	-123.440679572952 \\
262	-123.440679572952 \\
263	-123.440679572952 \\
264	-123.440679572952 \\
265	-123.440679572952 \\
266	-123.440679572952 \\
267	-123.440679572952 \\
268	-123.440679572952 \\
269	-123.440679572952 \\
270	-123.440679572952 \\
271	-123.440679572952 \\
272	-123.440679572952 \\
273	-123.440679572952 \\
274	-123.440679572952 \\
275	-123.440679572952 \\
276	-123.440679572952 \\
277	-123.440679572952 \\
278	-123.440679572952 \\
279	-123.440679572952 \\
280	-123.440679572952 \\
281	-123.440679572952 \\
282	-123.440679572952 \\
283	-123.440679572952 \\
284	-123.440679572952 \\
285	-123.440679572952 \\
286	-123.440679572952 \\
287	-123.440679572952 \\
288	-120.878850198848 \\
289	-120.878850198848 \\
290	-120.878850198848 \\
291	-120.878850198848 \\
292	-120.878850198848 \\
293	-120.878850198848 \\
294	-120.878850198848 \\
295	-120.878850198848 \\
296	-120.878850198848 \\
297	-120.878850198848 \\
298	-120.878850198848 \\
299	-120.878850198848 \\
300	-120.878850198848 \\
301	-120.878850198848 \\
302	-120.878850198848 \\
303	-120.878850198848 \\
304	-120.878850198848 \\
305	-120.878850198848 \\
306	-120.878850198848 \\
307	-120.878850198848 \\
308	-120.878850198848 \\
309	-120.878850198848 \\
310	-120.878850198848 \\
311	-120.878850198848 \\
312	-120.878850198848 \\
313	-120.878850198848 \\
314	-120.878850198848 \\
315	-120.878850198848 \\
316	-120.878850198848 \\
317	-120.878850198848 \\
318	-120.878850198848 \\
319	-120.878850198848 \\
320	-118.206167232445 \\
321	-118.206167232445 \\
322	-118.206167232445 \\
323	-118.206167232445 \\
324	-118.206167232445 \\
325	-118.206167232445 \\
326	-118.206167232445 \\
327	-118.206167232445 \\
328	-118.206167232445 \\
329	-118.206167232445 \\
330	-118.206167232445 \\
331	-118.206167232445 \\
332	-118.206167232445 \\
333	-118.206167232445 \\
334	-118.206167232445 \\
335	-118.206167232445 \\
336	-118.206167232445 \\
337	-118.206167232445 \\
338	-118.206167232445 \\
339	-118.206167232445 \\
340	-118.206167232445 \\
341	-118.206167232445 \\
342	-118.206167232445 \\
343	-118.206167232445 \\
344	-118.206167232445 \\
345	-118.206167232445 \\
346	-118.206167232445 \\
347	-118.206167232445 \\
348	-118.206167232445 \\
349	-118.206167232445 \\
350	-118.206167232445 \\
351	-118.206167232445 \\
352	-116.376780946552 \\
353	-116.376780946552 \\
354	-116.376780946552 \\
355	-116.376780946552 \\
356	-116.376780946552 \\
357	-116.376780946552 \\
358	-116.376780946552 \\
359	-116.376780946552 \\
360	-116.376780946552 \\
361	-116.376780946552 \\
362	-116.376780946552 \\
363	-116.376780946552 \\
364	-116.376780946552 \\
365	-116.376780946552 \\
366	-116.376780946552 \\
367	-116.376780946552 \\
368	-116.376780946552 \\
369	-116.376780946552 \\
370	-116.376780946552 \\
371	-116.376780946552 \\
372	-116.376780946552 \\
373	-116.376780946552 \\
374	-116.376780946552 \\
375	-116.376780946552 \\
376	-116.376780946552 \\
377	-116.376780946552 \\
378	-116.376780946552 \\
379	-116.376780946552 \\
380	-116.376780946552 \\
381	-116.376780946552 \\
382	-116.376780946552 \\
383	-116.376780946552 \\
384	-116.376780946552 \\
385	-116.376780946552 \\
386	-116.376780946552 \\
387	-116.376780946552 \\
388	-116.376780946552 \\
389	-116.376780946552 \\
390	-116.376780946552 \\
391	-116.376780946552 \\
392	-116.376780946552 \\
393	-116.376780946552 \\
394	-116.376780946552 \\
395	-116.376780946552 \\
396	-116.376780946552 \\
397	-116.376780946552 \\
398	-116.376780946552 \\
399	-116.376780946552 \\
400	-116.376780946552 \\
401	-116.376780946552 \\
402	-116.376780946552 \\
403	-116.376780946552 \\
404	-116.376780946552 \\
405	-116.376780946552 \\
406	-116.376780946552 \\
407	-116.376780946552 \\
408	-116.376780946552 \\
409	-116.376780946552 \\
410	-116.376780946552 \\
411	-116.376780946552 \\
412	-116.376780946552 \\
413	-116.376780946552 \\
414	-116.376780946552 \\
415	-116.376780946552 \\
416	-114.582657425991 \\
417	-114.582657425991 \\
418	-114.582657425991 \\
419	-114.582657425991 \\
420	-114.582657425991 \\
421	-114.582657425991 \\
422	-114.582657425991 \\
423	-114.582657425991 \\
424	-114.582657425991 \\
425	-114.582657425991 \\
426	-114.582657425991 \\
427	-114.582657425991 \\
428	-114.582657425991 \\
429	-114.582657425991 \\
430	-114.582657425991 \\
431	-114.582657425991 \\
432	-114.582657425991 \\
433	-114.582657425991 \\
434	-114.582657425991 \\
435	-114.582657425991 \\
436	-114.582657425991 \\
437	-114.582657425991 \\
438	-114.582657425991 \\
439	-114.582657425991 \\
440	-114.582657425991 \\
441	-114.582657425991 \\
442	-114.582657425991 \\
443	-114.582657425991 \\
444	-114.582657425991 \\
445	-114.582657425991 \\
446	-114.582657425991 \\
447	-114.582657425991 \\
448	-114.18531858492 \\
449	-114.18531858492 \\
450	-114.18531858492 \\
451	-114.18531858492 \\
452	-114.18531858492 \\
453	-114.18531858492 \\
454	-114.18531858492 \\
455	-114.18531858492 \\
456	-114.18531858492 \\
457	-114.18531858492 \\
458	-114.18531858492 \\
459	-114.18531858492 \\
460	-114.18531858492 \\
461	-114.18531858492 \\
462	-114.18531858492 \\
463	-114.18531858492 \\
464	-114.18531858492 \\
465	-114.18531858492 \\
466	-114.18531858492 \\
467	-114.18531858492 \\
468	-114.18531858492 \\
469	-114.18531858492 \\
470	-114.18531858492 \\
471	-114.18531858492 \\
472	-114.18531858492 \\
473	-114.18531858492 \\
474	-114.18531858492 \\
475	-114.18531858492 \\
476	-114.18531858492 \\
477	-114.18531858492 \\
478	-114.18531858492 \\
479	-114.18531858492 \\
480	-114.18531858492 \\
481	-114.18531858492 \\
482	-114.18531858492 \\
483	-114.18531858492 \\
484	-114.18531858492 \\
485	-114.18531858492 \\
486	-114.18531858492 \\
487	-114.18531858492 \\
488	-114.18531858492 \\
489	-114.18531858492 \\
490	-114.18531858492 \\
491	-114.18531858492 \\
492	-114.18531858492 \\
493	-114.18531858492 \\
494	-114.18531858492 \\
495	-114.18531858492 \\
496	-114.18531858492 \\
497	-114.18531858492 \\
498	-114.18531858492 \\
499	-114.18531858492 \\
500	-114.18531858492 \\
};
\end{axis}

\end{tikzpicture}

%% file: Images/table_summary.tex
\npdecimalsign{.}
\nprounddigits{1}
\sisetup{round-mode=places,round-precision=2}
\begin{tabular}{@{}lrrrrr@{}}
\toprule
      & {MCTS}            & {MLPDRL}             & {DRDRL} & {GRDRL Point} & {GRDRL Bin} \\ \midrule
{Average Collision Reward} & {\num{-192.9183048108}} & {\num{-229.7950990538}} & \num{-236.2505507}                       & \num{-148.48305716}                         & $\mathbf{-133.86} $                  \\ 
{Max Collision Reward}     & {\num{-145.7963114883}} & {\num{-139.379729308}}  & \num{-125.507133}                        & \num{-98.846263}                            & $\mathbf{-91.67}$                      \\
{Collisions Found}         & {21}              & {29}              & 30                                 & 25                                    & $\mathbf{32} $                                 \\ 
{Collision Percentage}     & \num{65.625}          & \num{90.625}          & \num{93.75}                              & \num{78.125}                                & $\mathbf{100}$                                 \\ \bottomrule
\end{tabular}
\npnoround